# Asymmetric Scalable Cross-modal Hashing

Wenyun Li
University of Macau
Macau, China

Chi-Man Pun*
University of Macau
Macau, China
cmpun@umac.mo

## ABSTRACT

Cross-modal hashing is a successful method to solve large-scale multimedia retrieval issue. A lot of matrix factorization-based hashing methods are proposed. However, the existing methods still struggle with a few problems, such as how to generate the binary codes efficiently rather than directly relax them to continuity. In addition, most of the existing methods choose to use an $n \times n$ similarity matrix for optimization, which makes the memory and computation unaffordable. In this paper we propose a novel Asymmetric Scalable Cross-Modal Hashing (ASCMH) to address these issues. It firstly introduces a collective matrix factorization to learn a common latent space from the kernelized features of different modalities, and then transforms the similarity matrix optimization to a distance-distance difference problem minimization with the help of semantic labels and common latent space. Hence, the computational complexity of the $n \times n$ asymmetric optimization is relieved. In the generation of hash codes we also employ an orthogonal constraint of label information, which is indispensable for search accuracy. So the redundancy of computation can be much reduced. For efficient optimization and scalable to large-scale datasets, we adopt the two-step approach rather than optimizing simultaneously. Extensive experiments on three benchmark datasets: Wiki, MIRFlickr-25K, and NUS-WIDE, demonstrate that our ASCMH outperforms the state-of-the-art cross-modal hashing methods in terms of accuracy and efficiency. The code will be available.

---

*Corresponding author

## 1 INTRODUCTION

Due to the development of the Internet, people generate a huge amount of multimedia data such as texts, videos, audio,and images. How to process this enormous data is a big issue in education[4], social media[40] ,and search technology[9]. And search and retrieval are very basic tasks in the multimedia data processing. In their solution, approximate nearest neighbor(ANN) search methods have gained a lot of attention. As a special method in ANN, hashing method first designed for uni-modal retrieval, which means the query and the train data are in the same modality[11],[20],[17]. However, for a more wide situation, people need a method that could work for cross-modal retrieval. Because in this case, the different dimensions would result in difficulty in searching and retrieval, it is attracting more attention, and cross-modal hashing[19] is a practical and powerful method that aims to generate a hash code to map the high-dimensional data to compact and comparable binary code. It can search for similarities in heterogeneous modalities, which has been highly concerned in media retrieval.And hash retrieval is very efficient due to its low storage requirement. In rough we classify cross-modal hashing methods into two classes, unsupervised cross-modal hashing and supervised cross-modal hashing.

However, there are several problems with these methods which need to be considered:

(1) Most supervised hashing methods simply utilize the inner product of label information to construct a semantic similarity affinity matrix, and for a large-scale dataset, because of the use of an $n \times n$ pairwise similarity matrix, the memory and the computational cost will be affordable.
(2) Due to the hash code is a discrete optimization problem that is very hard to solve, many methods relax the discrete hash code to continuous, however, this would lead to large quantization error and low-quality hash codes.
(3) Some hashing methods learn the hash codes and the hash function simultaneously, which would cause more complexity.

To address these challenges simultaneously, we propose a novel supervised cross-modal hashing method named Asymmetric Scalable Cross-Modal Hashing (ASCMH). The overview of our method is shown as Fig 1. We employ the collective matrix factorization to learn the common semantic latent space from different modalities. To preserve pairwise similarity while avoiding the large $n \times n$ similarity matrix, we adopt the distance-distance difference minimization method, so that the time and space complexity can be significantly reduced. We also utilize the orthogonalization quantization between the semantic labels and the generated hashing codes to fully exploit the label information. It is worth noting that the hash codes in the proposed ASCMH are generated discretely instead of by a relaxation scheme. Also the ASCMH is a two-step approach that could switch other powerful models as the proper hashing function to strengthen the discriminating power.

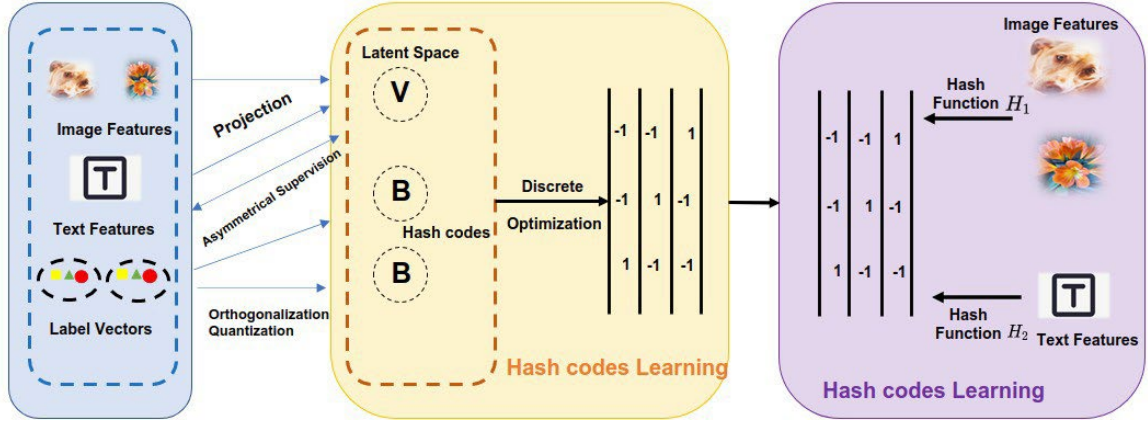

Figure 1: The framework of the proposed Asymmetric Scalable Cross-Modal Hashing (ASCMH).

The main contributions of this paper are summarized as the follows:

(1) A novel supervised cross-modal hashing method using Asymmetric Scalable Cross-Modal Hashing (ASCMH) is proposed. It can transform the discrete pairwise similarity matrix optimization into a distance-distance difference minimization problem. This makes our method salable to the large-scale datasets.
(2) A discrete optimization scheme for hashing codes is proposed in our ASCMH. The generation of binary hash codes without relaxation can avoid the large quantization error.
(3) Extensive experiments are conducted on three benchmark datasets: Wiki, MIRFlickr-25K, and NUS-WIDE. The results demonstrate that our method outperforms several state-of-the-art hashing methods for cross-modal retrieval.

## 2 RELATED WORK

As aforementioned, there are many hashing methods that have been proposed, which can be classified into unsupervised and supervised methods roughly. So we will briefly review some related works from these two perspectives.

### 2.1 Unsupervised Cross-Modal Hashing

For unsupervised cross-modal hashing methods, do not rely on the label information or semantic matrix and can obtain the hashing codes or build hashing functions by only exploring the relation of train data. There are a lot of unsupervised cross-modal hashing methods such as CMFH[5] which is using collective matrix factorization to get the hashing codes from the common latent semantic space of cross modalities; DRMFH[37] which ideally utilizes a unified objection function to get the consistency and inconsistency from different modalities;LSSH[43]which is utilizing spare coding and matrix factorization simultaneously to merge multiple latent semantic information to discriminate unified binary codes;FSH[16] uses the similarity fusion from different modalities to cope with cross-modality retrieval task.

### 2.2 Supervised Cross-Modal Hashing

In contrast, supervised cross-modal hashing methods[27][28][1, 2] take the advantage of semantic labels from multi-modality data points as the supervision to guide the learning of hashing codes or the generation of hashing functions.A huge amount of methods are proposed such as SMH[39] which can learn the hash functions bit by bit under the supervision of the label information; SCMH-WR[7] which can generate supervised the binary code directly without relaxing binary constraints, and more recently, due to the popularity of deep learning, e.g., DCMH[10] also achieve remarkable performance that thanks to the deep neural networks. In our work, we will be based on the supervised hash methods.

### 2.3 Deep Cross-Modal Hashing

Recently a lot of deep learning-based cross-modal hashing methods[6, 10, 13, 15, 30, 33, 34, 38][29, 35] are proposed, unlike other shallow hash methods using the hand-craft feature information as the input data, furthermore, deep neural networks have a strong ability to fit the nonlinear representation, and they can achieve more promising performance than shallow ones. For example, in [6], a two-step method are proposed, which includes features learning step and hashing generating step, and it takes advantage of the non-linear modeling ability of deep learning network, it learns the relations between the image-text pairs and modeling the special neural networks via probability. [38] proposed a model which combines the hashing encoding and classification can easily discriminate the data pair and preserve the semantic information. However, there are still some problems such as the time and computation consumption in deep learning training being too huge to afford in a large scale dataset, and the storing of the well-training model does not suit all cases in practice.

## 3 METHODOLOGY

### 3.1 Notations

In assumption, we define the training dataset consisting of $n$ instances, from multi-modality as $X^{(t)} = \{x_i^{(t)}\}_{i=1}^n$, where $x_i^{(t)} \in R^{d_t}$ will be defined as the $i$-th training data point from the $t$-th modality and $d_t$ is the feature dimension. Without loss of generality, we consider two modalities of data $X$ and $Y$. Moreover, in our model, $L = \{l_i\}_{i=1}^n \in \{0,1\}^{c \times n}$ is the label matrix, while $l_{ki} = 1$ means the data point $x_i$ is cataloged to the class $k$ or 0 otherwise. Our ASCMH's goal is attaining the unified hashing codes from training data points and modality-specific hashing generation functions for cross-modal retrieval of testing data points. The Frobenius norm of the matrix for $X$ is defined as $\|X\|_F^2 = tr(X^T X) = tr(XX^T)$, where $tr()$ means the trace operator. In addition, all training data are centralized in our experiments. So we can assume all data are zero-centered, i.e., $\sum_{i=1}^n x_i = 0$, and then all training data will be mapped to a $K$-dimensional kernel feature space with RBF kernel function, i.e., $\phi_i(x) = exp\left(-\frac{\|x - o_i\|^2}{2\sigma^2}\right)$, $\{o_i\}^k$ are the $k$ anchor points randomly selected from training data points, and $\sigma$ is the kernel width.

### 3.2 The Proposed Model

*3.2.1 Collective matrix factorization.* Matrix factorization is a powerful tool for information retrieval, as mentioned forefront, we assume in cross-modal data have consistent and inconsistent components, and the heterogeneous in multi-modality especially inconsistency has a huge impact in build the common latent semantic representation, which will lead to the loss and inaccuracy of cross-modal retrieval. So we aim at learning the consistency and inconsistency of different modalities.

So we adopt the collective matrix factorization technique[5] to extract the common semantic latent space from different modalities. So latent space learning is like the following:

$$\min_V \sum_r \lambda_c \|\phi\left(X^{(r)}\right) - P^{(r)}V\| \qquad (1)$$

where $\sum_r^l \lambda_c = 1$ is the hyper-parameter for balance. $P^{(r)} \in \mathfrak{R}^{k_r \times r}$ is the projection matrix, and $V \in \mathfrak{R}^{r \times n}$ is the desired common semantic latent space.

*3.2.2 Distance-Distance Difference Minimization.* Many supervised methods construct the inner-product similarity matrix to make use of the labels, which would lose precious information, especially for the multi-label data. What's worse, the pairwise similarity matrix for large-scale dataset will greatly increase. In order to solve this problem, we transform these problems by solving a distance-distance difference minimization problem as inspired by [12] and [32]. It means the distance difference between label pairwise and hash code pairwise shall be very small. So we have the following:

$$\min_B \sum_{i,j} \left(\|B_i - B_j\|_F^2 - r\|G_i - G_j\|_F^2\right) \qquad (2)$$
$$s.t. B \in \{-1, 1\}^{r \times n}$$

where G is a 2-norm normalized label matrix and defined as $G_i = L_i/\|L_i\|$. And we can easily have $\|G_i\|^2 = 1$ and $\|B\|^2 = r$ due to their propriety. Then we turn Eqn.2 as following:

$$\min_B \sum_{i,j}^n \left(\|B_i - B_j\|_F^2 - r\|G_i - G_j\|_F^2\right) \qquad (3)$$
$$\|B^T B - rG^T G\|_F^2, s.t. B \in \{-1,1\}^{r \times n}$$

In Eqn. 3, semantic information transforms to inner product semantic information search. More importantly, it could avoid the $n \times n$ size pairwise similarity matrix, which can apparently reduce the computation and memory cost.

Then we have learned the common semantic latent space $V$ with Eqn. 1 from different modalities. Then we need to explore the connection $V$ and the hash code $B$. Here we adopt the similar method in [41], and define the following:

$$B = VR \qquad (4)$$
$$s.t. R^T R = I, VV^T = nI_r, V1_n = 0_r$$

where $R \in \mathfrak{R}^{r \times n}$ is the connection matrix between $B$ and $V$, and the orthogonal constraint of $R$ makes each column of $R$ independent of each other.

What's more, only depending on pairwise similarity matrix is not enough. For example, supposing there are three instances with label $l_1 = [1, 0, 0]$, $l_2 = [1, 0, 0]$ and $l_3 = [1, 0, 0]$, while the pairwise similarity metrics between $l_1$ to $l_2$ and $l_2$ to $l_3$ are both 1. However it is obvious that $l_2$ is more similar to $l_3$ than $l_1$. So we have enough necessity to embed the label matrix into the learning of hash codes, i.e.,

$$\min_{B,M} \|B - L\|^2 \qquad (5)$$
$$s.t. B \in \{-1, 1\}^{r \times n}$$

where $M \in \mathfrak{R}^{c \times r}$ is the projection from $L$ to $B$. So we reformulate Eqn.3 in the condition of Eqn.4 and Eqn.5 into the following:

$$\min_{V,R,M} \|VR(LM)^T - G^T G\|_F^2 + \|B - LM\|_F^2 \qquad (6)$$
$$s.t. B \in \{-1, 1\}^{r \times n}, R^T R = I$$

*3.2.3 Overall Objective Function.* Combining Eqn. 1, and Eqn. 6 together, we have the following objective function of our method:

$$\min_{V,R,M,B,P^{(r)}} \|VR(LM)^T - rG^T G\|_F^2 + \omega \|B - LM\|_F$$
$$\sum_{r=0}^l \lambda_c \|\phi\left(X^{(r)}\right) - P^{(r)}V\|_F^2 \qquad (7)$$
$$s.t. B \in \{-1,1\}^{r \times n}, R^T R = I, VV^T = nI_r, V1_n = 0_r$$

where $\omega$ and $\lambda_c$ are the hyper-parameters.

## 3.3 Optimization

Directly solving the Eqn. 7 may be very difficult since Eqn. 7 contains the discrete value and it is an NP problem. So in order to tackle this optimization problem, we proposed an efficient alternative iterative optimization algorithm. At each step, we solve the optimization by only updating one variable while maintaining the others fixed and repeating the procedure until convergence.

(1) Optimizing $P^{(r)}$.

When other variables are fixed, Eqn. 7 is then simplified as following formulation:

$$\min_{P^{(r)}} \|\phi\left(X^{(r)}\right) - P^{(r)}V\|_F^2 \qquad (8)$$

By taking the partial derivative of Eqn. 7 with respect to $P^{(r)}$ and setting it to zero, $P^{(r)}$ can be updated by:

$$P^{(r)} = \phi\left(X^{(r)}\right) V^T (VV^T)^{-1} \qquad (9)$$

Due to the condition of $VV^T = nI_r$, the formula of $P^{(r)}$ is equivalent to,

$$P^{(r)} = \phi\left(X^{(r)}\right) V^T / n \qquad (10)$$

(2) Optimizing $M$.

Given the other variables, the subproblem in Eqn. 7 with respect to $M$ reduces to the following formulation

$$\min_{M} \|VR(LM)^T - rG^TG\|_F^2 + \|B - LM\|_F \qquad (11)$$

By taking the partial derivative of function 11 with respect to $M$ and setting it to zero, $M$ can be updated by:

$$M = (\mu + 1) L^T L^{-1} \left(rL^T GG^T VR + \mu L^T B\right) \qquad (12)$$

(3) Optimizing $R$.

Given the other variables, the subproblem in Eqn. 7 with respect to $R$ reduces to the following formulation

$$\min_{R} \|VR(LM)^T - rG^TG\|_F^2 \qquad (13)$$
$$s.t. R^T R = I$$

Because of Eqn. 13 is an orthogonal rotation problem, so it's hard to optimize, we develop an alternative version of 13, we directly maximize the similarity matrix and the inner product of asymmetric hashing codes embedding with $V^T V = I_r$, i.e.,

$$\max_{R} Tr\left(rM^T L^T GG^T R\right) \qquad (14)$$
$$s.t. R^T R = I$$

Obviously, the current model w.r.t. $R$ is a classic Orthogonal Procrustes Problem (OPP), and it can be solved by using simple SVD(singular value decomposition). Let $P_l$ and $P_r$ be the left and right singular matrices of $rM^T L^T GG^T$, i.e.

$$[P^l, \Sigma, P^r] = svd(r^T M^T L \, G^T G) \qquad (15)$$

Then we can get the optimal solution of $R$ is

$$R = P^l (P^r) \qquad (16)$$

(4) Optimizing $V$.

Given the other variables, the subproblem in Eqn. 7 with respect to $V$ reduces to the following formulation

$$\min_{V} \|VR(LM)^T - rG^TG\|_F^2 + \sum_r \lambda_c \|\phi\left(X^{(r)}\right) - P^{(r)}V\|_F^2 \qquad (17)$$
$$s.t. VV^T = nI_r$$

To solve the above problem, we transform Eqn. 17 into a matrix trace form with the constrains of $VV^T = nI_r$ as the following:

$$\max_{V} Tr\left(rLMR^T V^T GG^T + \sum_{r=0}^{l} \lambda_c P^{(r)} \phi\left(X^{(r)}\right) V^T\right) \qquad (18)$$
$$s.t. V^T V = I$$

To simplify the representation, we define $J = I_n - \frac{1}{n} 1_n 1_n^T$ and $Z = rGG^T LMR^T + P^{(r)} \phi\left(X^{(r)}\right)$. So with the known constrains, we perform the eigen decomposition of $ZJZ^T$ like,

$$ZJZ^T = [Q \; \bar{Q}] \begin{bmatrix} \Omega & 0 \\ 0 & 0 \end{bmatrix} [Q \; \bar{Q}]^T \qquad (19)$$

where $\Omega \in \mathcal{R}^{r' \times r'}$ and $Q \in \mathcal{R}^{K \times r'}$ are the diagonal matrices of the positive eigenvalues and the corresponding eigenvectors, while $\bar{Q}$ is the matrix of the remaining $r - r'$ eigenvectors, and we can easily get the orthogonal matrix $\bar{Q} \in \mathcal{R}^{r' \times (r-r')}$. Then we define $\bar{P} = JZ^T Q\Omega^{-\frac{1}{2}}$ and a random orthogonal matrix $\bar{P} \in \mathcal{R}^{n \times (r-r')}$. According to [18], we get the optimal solution of Eqn. 18 as,

$$V = \sqrt{n} [Q \; \bar{Q}] [\bar{P} \; \bar{P}]^T \qquad (20)$$

(5) Optimizing $B$.

Given the other variables, the subproblem in function 7 with respect to $B$ reduces to the following formulation

$$\min_{B} \|B - LM\|_F^2 \qquad (21)$$
$$s.t. B \in \{-1, 1\}^{r \times n}$$

Here we take the same action to $V$. Finally, we have the optimal solution,

$$B = sgn(LM) \qquad (22)$$

which the definition of $sgn()$ is the element-wise indicator operator.

In order to obtain the final solution, we can update alternately optimized variables, i.e., $P^{(r)}, R, V, M, B$ according to the above steps until it converges or reaches the maximum iterations. To demonstrate the above learning algorithm clearly, we summarize it in Algorithm 1

**Algorithm 1:** Optimization algorithm in ASCMH

**Input:** Feature matrix of image-text pairs $O$ and Label matrix L;Parameters $\lambda,\alpha,\beta,\gamma$ and $\mu$.
**Output:** Optimal $B$.
1 Mapping the image-text pairs to kernel spaces to $X,Y$.
2 Randomly initialize the projection matrix $M$,the orthogonal rotation projection matrices $R$ and the common space matrix $V$, desired hash codes matrix $B$,the projection matrix $P^{(r)}$.
3 **while** *convergence criterion NOT satisfied* **do**
4     $P^{(r)}$-step:Update $P^{(r)}$ according 10
5     $M$-step:Update $M$ according 12
6     $R$-step:Update $R$ according 16
7     $V$-step:Update $V$ according 20
8     $B$-step:Update $B$ according 22

### 3.4 Hash Functions Learning

Many previous works [21, 22] have shown that two-step hash functions are vital for performance. Hence we choose to leverage the two-step hashing function in our model. Instead of any classification functions such as deep learning classifier[31, 42], here we choose the linear regression on account of its simplicity, the projection matrix $P_h$ which maps the training data points $X$ to $B$, and the two-step hash optimizing functions with a regularization term:

$$\|B - XP_h\|_F^2 + \lambda_h \|P_h\|_F \quad (23)$$

where $\lambda_h$ is a balance parameter. The optimal $P$ can be expressed as,

$$P_h = (X^T X + \lambda_h I)^{-1} X^T B \quad (24)$$

So the two step ASCMH as,

$$B_{ts} = sgn(XP_h) \quad (25)$$

### 3.5 Complexity Analysis

The computational complexity of the proposed ASCMH is discussed as following. For each iteration, the computational complexity for learning the Eqn. 1 is $O\left(\sum_{l=1}^{r_m}(d_l k_l)n\right)$. The complexity of updating $P^{(r)},M,R,V,B$ are $O\left(T\left(\sum^{r_m} k_l rn\right)\right),O\left(T\left(r^3 + r^2 n + rn\right)\right),$ $O\left(T\left(r^3 + r^2 n\right)\right), r^3+r^2+2*r^n+rn+P^{(r)}, O\left(\sum_{l=1}^{r_m}(d_l k_l)n\right)$ relatively, where $n$ is the size of the training set, $d_l$ is the original feature dimensionality, $k_l$ is the number of anchors after kernelization. $k_0$ is the number of label categories, $r$ is the hash code length, $T$ is the number of iterations. Due to $T, d_l, k_l, k_0, \ll n$, we can draw the conclusion that our scheme's complexity of the training is linear to the size of the training set $n$. So our ASCMH is suitable for large-scale datasets.

## 4 EXPERIMENTS

### 4.1 Datasets

**Wiki**[25] is the most common benchmark dataset for cross-modal retrieval. There are 10 most populated categories which are considered. There are 2,866 image/text pairs training data, all image and text data are turning to 128-dimensional SIFT feature and 10-dimensional topic vector.

**NUS-WIDE**[3] The NUS-WIDE dataset contains 269,648 images with a total of 5,018 tags collected from the National University of Singapore, which is categorized into 81 groups. Finally, 5,018 unique tags have corresponded with images, and each image is associated with 6 tags on average.

**MIRFlickr25K**[8] Flickr-25K has 2,5000 pictures, and each picture has corresponding tags and annotations. Tags can be used as text descriptions, among which 1,386 tags appear in at least 20 pictures; There are 24 annotations as labels.

### 4.2 Metrics

We conducted two cross-modal retrieval tasks including Image-to-Text which uses images as the query to search texts and Text-to-Image which uses texts as the query to search images. Just as in other work, we adopt mean Average Precision(mAP) as the evaluation metric for effectiveness in our experiment. Given a query and a set of R retrieved instances, the Average Precision(AP) is defined as

$$AP = \frac{1}{L} \sum_{r=1}^{R} P_r \delta(r) \quad (26)$$

where $L$ is the number of ground-truth neighbors in the retrieval set, $R$ is the size of the retrieval set, $P_r$ denotes the precision of the top $r$ retrieved instances, and $\delta_r$ is an indicator function which equals to 1 if the $rth$ instance is relevant to query or 0 otherwise. Then the AP of all queries is averaged to obtain the mAP as the following definition,

$$mAP = \frac{1}{L_q} \sum_{i=1}^{L_q} AP(i) \quad (27)$$

where $L_q$ is the size of the query set. From the definition we know the larger the MAP is, the better the performance is.

### 4.3 Baselines

In our experiment, we compare our ASCMH method with several state-of-the-art shallow hash methods including CMFH[5], LSSH[43], DCH[36], FSH[16], SCRACH[14], GSPH[23], BATCH[32], SDDH[24] and SLCH[26]. Our most baseline methods are using from their author's paper with their suggestion parameters. In our study, we conducted a parameter sensitivity analysis on MIRFlickr-25K, and part of the results are shown in the table. The proposed ASCMH achieves the best performance when $\omega = 0.01, \lambda_1 = 0.5$ and $\lambda_2 = 0.5$ on MIRFlickr-25K from the section 4.4.4. Just as in the other paper's setting, we followed their setting, all the results are averaged over 20 runs. The kernel feature dimensionality $k$ is set to 500 and 1,000 for image and text, respectively. All the experiments are performed on a computer with Intel Core i7-10700K CPU and 32 GB RAM.

### 4.4 Results

In this section, we conduct the two most common cross-modal retrieval tasks on the Image-query-Text (using an image to query similar texts) and Text-query-Image (using a text to query similar images). Just as same with the other paper, we adopt mean average precision and Top-N precision as the evaluation metrics. In our experiments, we set the code lengths to 16,32,64 , and 128.

*4.4.1 Results on Wiki.* The MAP results of ASCMH and the baselines on Wiki are shown in Table 2. It contains the Image-to-Text and the Text-to-Image tasks, and the code length varies from 16 to 128. The best results are shown in boldface. The top-N precision curves of the cases with 64 bits are plotted in Fig 4. From these results, we can find that: 1) ASCMH achieves the best MAP results compared with all baselines, both in Image-to-Text and Text-to-Image can achieve state of the art. 2) Compared with MIRFlickr-25K and NUS-WIDE MAP results, we can find that all methods are not good at Wiki dataset. The main reason is that the image and text features of the Wiki are much low-dimensional, so it is very hard to hash learning the information. 3) All methods achieve better performance with the code length increasing, as longer bits can carry more useful information.

*4.4.2 Results on MIRFlickr-25K.* Table 2 shows the mAP results for all mentioned methods with ours in different code lengths on three datasets, and Fig. 5 is the curves on MIRFlickr-25K. From the results, we can find the followings: 1) Our ASCMH method significantly outperforms nearly all the baselines when compared to other methods. 2) Compared with the unsupervised methods , the supervised methods can achieve better performance because they can find the connection to real-world label information. 3) With the code length increasing, the performance of ASCMH is growing, meaning the longer hash codes could carry more information from their modalities. 4) Compared with the task "Img to Txt", most methods generally achieve better performance on "Txt to Img", which may be due to the text data can search the demanded information better than image data.

*4.4.3 Results on NUS-WIDE.* Fig.6 shows the Top-N precision curves of previous compared methods in the condition of 64-bits hash code on NUS-wide. Table 2 is the MAP results of ASCMH and the baselines on NUS-wide. From the figure and table, we can learn that:1) our ASCMH is better than all methods even the recent method like BATCH[32]; 2) our method except in 32-bits, gains the best performance in NUS-wide; 3) Generally speaking, all methods achieve better performance with the code length increasing in the NUS-wide dataset. The reason is that the longer bits can carry more useful information.

*4.4.4 Parameter Analysis.* In order to analyze the sensitivity of parameters including $\omega$ and $\lambda_c$ in 7, we conducted an experiment. In specific, we experiment by varying the value of one parameter while fixing the others on MIRFlickr-25K dataset on the 32-bits. Here, we vary $\omega$ from 0 to 1 while fixing $\lambda_1$ and $\lambda_2$. The results on both the Image-to-Text and the Text-to-Image tasks are shown in Fig 2. So we can observe that $\omega$ is not a significant parameter in the retrieval processing, and generally $\omega$ will affect the Image-to-Text and the Text-to-Image tasks both. And from the result, we can find that when $\omega = 0.5$ two tasks will have higher scores while balancing both tasks simultaneously.

*4.4.5 Time Cost Analysis.* In section 3.5, we analyze the computational complexity of ASCMH and show that it is linear to the size of the training set. Here we compared the training time of ASCMH with several supervised baseline shallow methods. The experiments are conducted on MIRFlickr25k by varying the training set size from 3k to 20k. The results are shown in Table1, which is clear that the training time is linear to the dataset size.

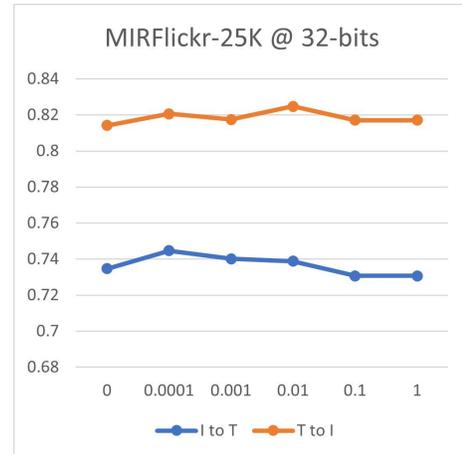

**Figure 2: Sensitivity analysis of parameter $\omega$ on MIRFlickr-25K.**

## 5 CONCLUSIONS

In this work, we present a novel supervised cross-modal hashing method using Asymmetric Scalable Cross-Modal Hashing (ASCMH). It can explore consistency and inconsistency by learning across different modalities. We also propose a discrete optimization scheme for hashing codes in our ASCMH, so that the generation of binary hash codes without relaxation can avoid the large quantization error. Extensive experiments demonstrate the effectiveness of our method in cross-modal retrieval in comparison with other state-of-the-art cross-modal retrieval methods on three common benchmark datasets. However, the hashing method is hard to handle some extreme non-linear problem, and lack the ability of digging the relation of high level and low level semantic information. In future work, it is possible to add some activation function in the modeling. Besides, some data augmentation will improve the effectiveness in the training.

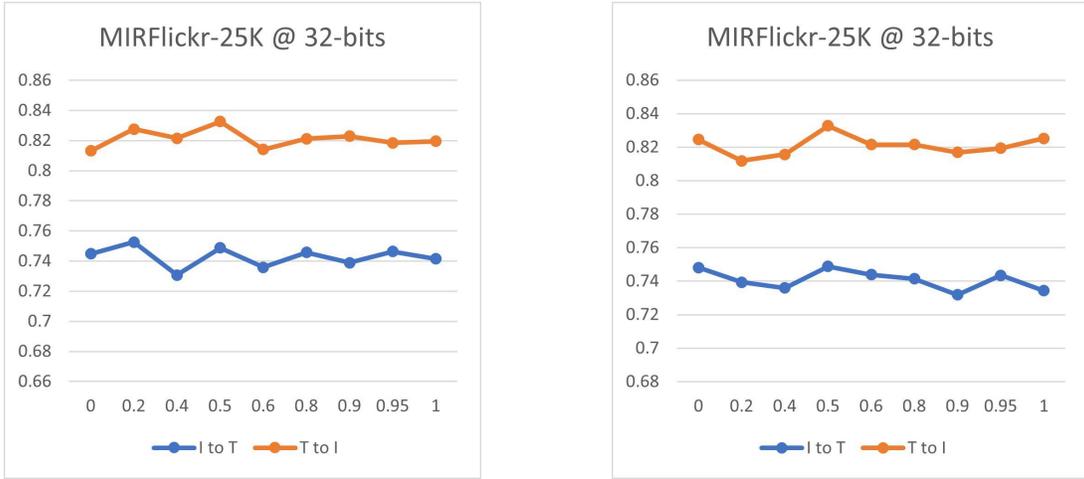

Figure 3: Sensitivity analysis of parameter $\lambda_1$ and $\lambda_2$ on MIRFlickr-25K.

Table 1: The training time (second) of ASCMH on NUS-WIDE with different size

| Size | 8-bits | 16-bits | 24-bits | 32-bits | 64-bits | 96-bits | 128-bits |
|---|---|---|---|---|---|---|---|
| 3k | 0.287 | 0.299 | 0.299 | 0.310 | 0.352 | 0.394 | 0.441 |
| 5k | 0.663 | 0.684 | 0.689 | 0.703 | 0.746 | 0.836 | 0.911 |
| 6k | 0.909 | 0.924 | 0.928 | 0.947 | 1.034 | 1.100 | 1.219 |
| 9k | 1.899 | 1.948 | 1.916 | 1.933 | 2.002 | 2.182 | 2.306 |
| 10k | 2.251 | 2.293 | 2.227 | 2.283 | 2.426 | 2.537 | 2.729 |
| 20k | 7.702 | 7.662 | 7.659 | 7.850 | 8.130 | 8.329 | 8.558 |

Table 2: The mAP values on three datasets for various code lengths

| Task | Method | Wiki | | | | MIRFlickr25k | | | | NUS-wide | | | |
|---|---|---|---|---|---|---|---|---|---|---|---|---|---|
| | | 16 bits | 32 bits | 64 bits | 128 bits | 16 bits | 32 bits | 64 bits | 128 bits | 16 bits | 32 bits | 32 bits | 128 bits |
| Img to Text | CMFH | 0.1664 | 0.165 | 0.1669 | 0.1739 | 0.399 | 0.4063 | 0.3975 | 0.3962 | 0.6503 | 0.6516 | 0.6531 | 0.6486 |
| | LSSH | 0.2131 | 0.2177 | 0.2144 | 0.228 | 0.3887 | 0.3848 | 0.3874 | 0.3896 | 0.6392 | 0.6644 | 0.6586 | 0.6682 |
| | DCH | 0.3192 | 0.3377 | 0.3535 | 0.3722 | 0.5595 | 0.5793 | 0.5944 | 0.6015 | 0.7176 | 0.724 | 0.7285 | 0.7314 |
| | FSH | 0.2357 | 0.251 | 0.2524 | 0.2513 | 0.3578 | 0.3826 | 0.3912 | 0.3994 | 0.6457 | 0.6594 | 0.6618 | 0.6712 |
| | SCRACH | 0.3428 | 0.3497 | 0.3662 | 0.3613 | 0.7065 | 0.7045 | 0.7161 | 0.7248 | 0.6072 | 0.6332 | 0.6456 | 0.6474 |
| | GSPH | 0.2392 | 0.2772 | 0.2965 | 0.3018 | 0.7159 | 0.7346 | 0.7413 | 0.7453 | 0.5489 | 0.5725 | 0.5832 | 0.5887 |
| | BATCH | 0.3553 | 0.3781 | 0.3837 | 0.3940 | 0.7318 | **0.7408** | 0.7416 | 0.7464 | 0.6274 | 0.6505 | 0.6632 | 0.6638 |
| | SDDH | 0.3544 | 0.3739 | 0.3857 | 0.3904 | 0.721 | 0.7394 | **0.7454** | 0.7494 | 0.6504 | 0.6564 | **0.6670** | 0.6633 |
| | SLCH | 0.3526 | 0.3648 | 0.3537 | 0.3829 | 0.6294 | 0.6238 | 0.6269 | 0.6472 | 0.6161 | 0.6036 | 0.6367 | 0.6425 |
| | **ASCMH** | **0.3664** | **0.3948** | **0.3911** | **0.3950** | **0.7319** | 0.7371 | 0.7367 | **0.7523** | **0.6510** | **0.6588** | 0.6633 | **0.6638** |
| Text to Img | CMFH | 0.4563 | 0.4751 | 0.4986 | 0.5299 | 0.4255 | 0.4278 | 0.4215 | 0.4187 | 0.3498 | 0.3435 | 0.3486 | 0.3529 |
| | LSSH | 0.494 | 0.5072 | 0.5288 | 0.5402 | 0.4101 | 0.413 | 0.4176 | 0.4151 | 0.4219 | 0.4228 | 0.4249 | 0.4192 |
| | DCH | 0.6724 | 0.7097 | 0.7216 | 0.7141 | 0.667 | 0.7083 | 0.7285 | 0.7390 | 0.7106 | 0.7103 | 0.7260 | 0.7223 |
| | FSH | 0.4092 | 0.4864 | 0.4961 | 0.5247 | 0.5869 | 0.5979 | 0.6186 | 0.6251 | 0.4334 | 0.4446 | 0.4714 | 0.4819 |
| | SCRACH | 0.7398 | 0.7538 | 0.7496 | 0.7513 | 0.7685 | 0.7773 | 0.7936 | 0.8007 | 0.7488 | 0.7747 | 0.7855 | 0.7893 |
| | GSPH | 0.2467 | 0.2690 | 0.2829 | 0.2823 | 0.7405 | 0.7567 | 0.7615 | 0.7705 | 0.6779 | 0.6893 | 0.6921 | 0.6962 |
| | BATCH | 0.7382 | 0.7448 | 0.7586 | 0.7502 | 0.8005 | **0.8187** | 0.8192 | 0.8256 | 0.7596 | 0.7782 | 0.7815 | 0.7837 |
| | SDDH | 0.7452 | 0.7554 | 0.7590 | 0.7618 | 0.7917 | 0.8132 | 0.8241 | 0.8328 | 0.7638 | 0.7790 | **0.7945** | 0.7980 |
| | SLCH | 0.7356 | 0.7251 | 0.7343 | 0.7354 | 0.6721 | 0.6682 | 0.6761 | 0.7064 | 0.7298 | 0.7281 | 0.7660 | 0.7678 |
| | **ASCMH** | **0.7481** | **0.7607** | **0.7590** | **0.7628** | **0.8062** | 0.8125 | **0.8286** | **0.8339** | **0.7677** | **0.7824** | 0.7898 | **0.7990** |

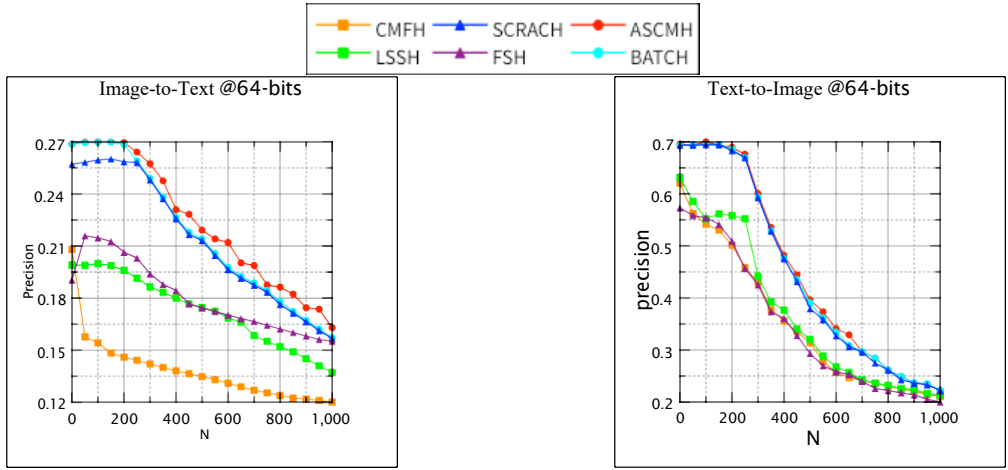

Figure 4: The top-N precision on wiki

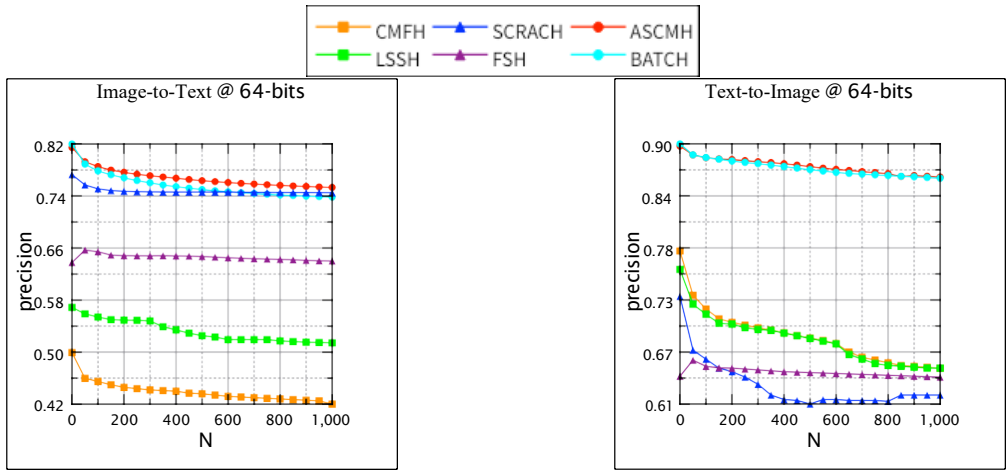

Figure 5: The top-N precision on MIRFlickr-25K

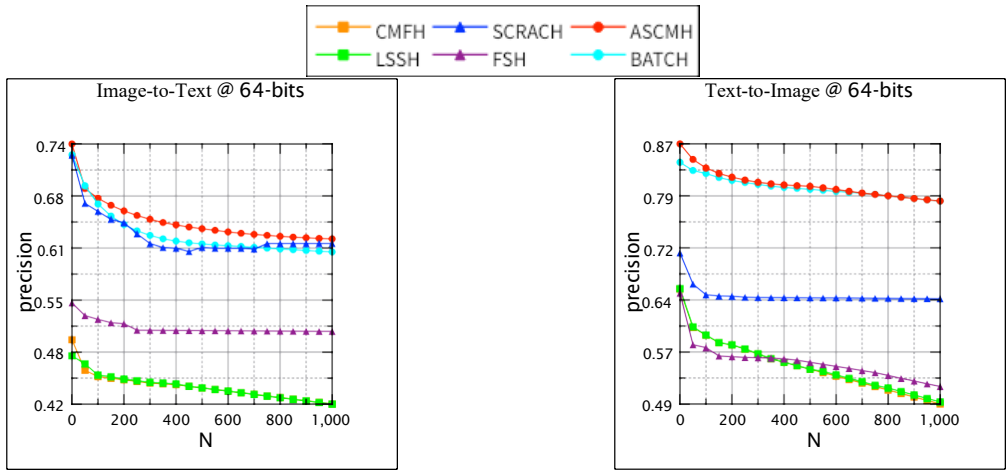

Figure 6: The top-N precision on nus-wide